\PassOptionsToPackage{table}{xcolor}
\documentclass[]{alaya}
\usepackage{makecell}
\usepackage{wrapfig}
\usepackage{tabularx}
\usepackage{textcomp}
\usepackage{stfloats}
\usepackage{url}
\usepackage{verbatim}
\usepackage{titlesec}
\usepackage{tocloft}
\usepackage{adjustbox}
\usepackage{multirow}
\usepackage{pifont}
\usepackage[sc]{mathpazo}
\usepackage{tikz}
\usepackage{comment}
\usepackage{amsmath,amssymb}
\usepackage{colortbl}
\usepackage{natbib}
\usepackage{color}
\usepackage{booktabs} 
\usepackage{hyperref}
\usepackage{graphicx}
\usepackage{subcaption}
\usepackage{multirow}
\usepackage{subcaption}
\RequirePackage{xspace}
\makeatletter
\DeclareRobustCommand\onedot{\futurelet\@let@token\@onedot}
\def\@onedot{\ifx\@let@token.\else.\null\fi\xspace}
\usepackage[most]{tcolorbox}
\usepackage{xcolor}
\usepackage{array}
\usepackage{tabularx}
\usepackage{siunitx} 
\usepackage{makecell}
\usepackage[table]{xcolor}
\usepackage{caption}
\definecolor{headerpurple}{HTML}{d8d2fc}
\definecolor{rowgray}{gray}{0.95}

\def\ie{\emph{i.e}\onedot}

\makeatother

\definecolor{adptorange}{RGB}{248, 205, 172}
\definecolor{cmpblue}{RGB}{189, 215, 238}
\definecolor{cmpblue}{RGB}{189, 215, 238}

\definecolor{our_red}{RGB}{232,157,160}
\definecolor{our_blue}{RGB}{136,206,230}
\definecolor{our_orange}{RGB}{246,200,168}
\definecolor{our_green}{RGB}{178,211,164}

\definecolor{attn_code0}{RGB}{247,215,200}
\definecolor{attn_code1}{RGB}{238,169,139}
\definecolor{mlp_code0}{RGB}{204,201,221}
\definecolor{mlp_code1}{RGB}{102,95,153}
\definecolor{mygray}{HTML}{f0f0f0}

\definecolor{token_blue}{RGB}{84, 120, 140}

\usepackage{pifont}
\usepackage{bbding}
\usepackage{fontawesome}
\usepackage{xspace}

\usepackage{float}

\newlength\savewidth

\newcolumntype{x}[1]{>{\centering\arraybackslash}p{#1pt}}
\newcolumntype{y}[1]{>{\raggedright\arraybackslash}p{#1pt}}
\newcolumntype{z}[1]{>{\raggedleft\arraybackslash}p{#1pt}}

\renewcommand{\paragraph}[1]{\vspace{1mm}\noindent\textbf{#1}}
\usepackage{colortbl}
\usepackage{xcolor}
\usepackage{wrapfig}

\renewcommand{\paragraph}[1]{\vspace{1.25mm}\noindent\textbf{#1}}

\usepackage{algorithm}
\usepackage{listings}

\definecolor{codeblue}{rgb}{0.25, 0.5, 0.5}
\definecolor{codekw}{rgb}{0.35, 0.35, 0.75}
\lstdefinestyle{Pytorch}{
    language = Python,
    backgroundcolor = \color{white},
    basicstyle = \fontsize{9pt}{8pt}\selectfont\ttfamily\bfseries,
    columns = fullflexible
    aboveskip=1pt,
    belowskip=1pt,
    breaklines = true,
    captionpos = b,
    commentstyle = \color{codeblue},
    keywordstyle = \color{codekw},
}

\definecolor{green}{HTML}{009000}
\definecolor{red}{HTML}{ea4335}

\title{WorldSculpt: Generating Compositional Worlds from Grounded Videos}

\author[\triangle,\clubsuit]{Muyao Niu}
\author[\triangle]{Jixuan He}
\author[\triangle]{Ruihan Yu}
\author[\triangle]{Lian Fu}
\author[\triangle]{Yonghao Yu}
\author[\triangle]{Zheng-Hui Huang}
\author[\triangle]{Yifan Zhan}
\author[\triangle]{Fengbo Lan}
\author[\triangle]{Yongtao Ge}
\author[\clubsuit]{Yinqiang Zheng}
\author[\triangle]{Kaipeng Zhang}
\author[\triangle]{Zhixiang Wang}

\affiliation[\triangle]{Alaya Lab}
\affiliation[\clubsuit]{The University of Tokyo}

\abstract{
We study the problem of generating a compositional 3D representation of a cluttered scene containing hundreds of objects. The goal is to represent the scene as a collection of individual object meshes placed in a shared world frame, as required by downstream applications such as gaming, AR/VR, simulation, and robotics. This task is challenging in densely cluttered scenes, where objects heavily occlude one another and each view reveals only a fraction of their geometry. Geometry-based approaches typically reconstruct the scene as a single representation and leave incomplete geometry in occluded regions, while existing compositional methods with generative priors are largely limited to relatively simple scenes. We show that complex scenes with hundreds of objects can instead be generated compositionally by adapting a strong single-object 3D generative prior to multi-view observations. We instantiate this paradigm with Pixal3D, extending it with a multi-view conditioning pathway that grounds object generation in multiple posed observations. Although the model is finetuned entirely on single objects in canonical space, it generalizes to large scenes with severe occlusion without any scene-level training, demonstrating the feasibility and scalability of this paradigm. We further introduce UE-MeshyScene, a photorealistic benchmark of densely cluttered scenes with hundreds of objects, per-object annotations, and ground-truth meshes. Across single-object, controlled multi-object, and UE-MeshyScene evaluations, our method consistently outperforms prior approaches, with larger gains as scene complexity and occlusion increase. Finally, we demonstrate broader applicability by converting generated 3DGS worlds, such as Marble and HY-World 2.0, into compositional mesh scenes.

}

\project{\url{https://alaya-lab.github.io/WorldSculpt}}
\code{\url{https://github.com/AlayaLab/WorldSculpt}}
\correspondence{Zhixiang Wang (Project Lead), Kaipeng Zhang}
\date{\today}

\begin{document}
\maketitle

\begin{figure}[!h]
\centering
  \includegraphics[width=0.99\linewidth]{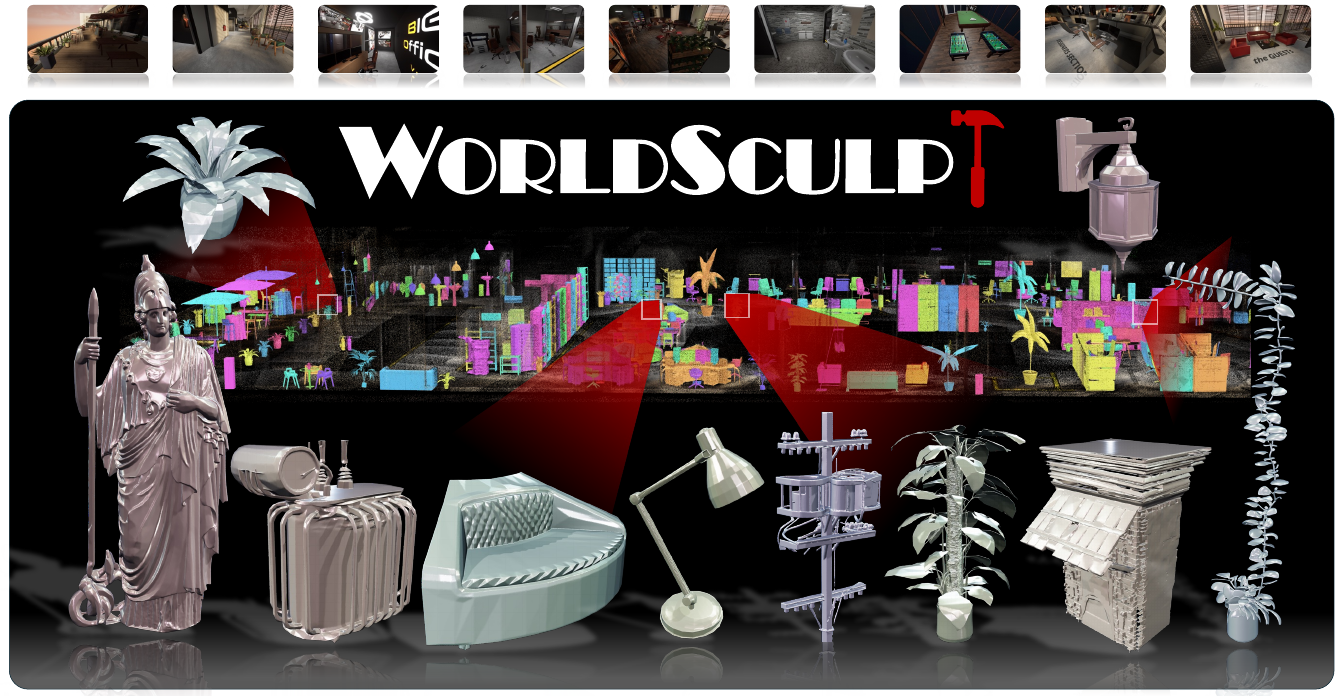}
  \setlength{\abovecaptionskip}{2mm}
  \captionof{figure}{\textbf{WorldSculpt.} Given a set of RGB images with instance-level 2D masks and 3D bounding boxes, WorldSculpt produces a compositional mesh representation for very complex scenes consisting of hundreds of individual objects.}
\label{fig:teaser}
\end{figure}

\section{Introduction}
\label{sec:intro}

Recent generative world models have demonstrated impressive capabilities in synthesizing persistent and explorable 3D environments from only a single image or a short text prompt. Systems such as Marble~\cite{worldlabs2025marble} and HY-World 2.0~\cite{team2026hy} can generate plausible geometry and appearance well beyond the directly observed content. However, the resulting world is typically represented as a unified scene representation, such as a fused mesh or a set of Gaussians, without explicitly separating individual objects. As a result, objects such as a chair or a vase cannot be independently selected, moved, or transferred into a physics simulator, since they are not represented as separate entities. This limitation creates a fundamental mismatch with the requirements of many downstream applications, including gaming and content creation, AR/VR, simulation, and robotics. These applications do not consume an undifferentiated soup of surface geometry. Instead, they require a scene to be decomposed into its constituent objects, with each represented by a mesh that can be independently selected, moved, and re-simulated.

Generating compositional 3D content from multiple images of a real cluttered scene remains a long-standing and largely unsolved problem in computer vision. In this setting, the structure of the output is just as important as its geometric accuracy. The task is particularly challenging in complex scenes, where many objects are densely packed and heavily occlude one another, leaving only partial observations of each object in any individual view. A faithful result must therefore align the visible evidence across images, recover plausible geometry for unobserved regions, and preserve each object as an independently usable asset.

Existing approaches address different aspects of this problem, but none fully satisfy all of these requirements. Geometry-centric methods, including per-scene optimization, feed-forward volumetric or Gaussian regressors, and pointmap foundation models~\cite{wang2024dust3r,wang2025vggt,lin2025depth}, can recover accurate geometry in observed regions. However, they typically reconstruct the scene as a single fused representation and leave missing geometry in areas hidden by occlusion. Generative priors offer a complementary strength. By learning a distribution over plausible 3D content, they can infer coherent geometry beyond the visible evidence, but existing methods generally fall into two categories, each with its own limitation. Scene-level world generation models such as Marble process the entire scene jointly and can synthesize unobserved regions, yet their output is still a single monolithic representation. Native-3D generative priors~\cite{xiang2025structured,zhao2025hunyuan3d,li2026pixal3d}, on the other hand, model complete individual objects, but are designed for a single pose-free image of one object. They do not jointly condition on a consistent set of views or generate objects directly in a known scene coordinate frame, making them difficult to apply to complex scenes without additional machinery. Recent compositional approaches~\cite{huang2025midi,meng2026scenegen,shi2026scenemaker,lin2026partcrafter} instead extend object-level generative priors to multi-object scenes, either by jointly generating multiple instances or by combining object generation with scene-level layout estimation. In practice, however, these methods have primarily been demonstrated on relatively simple scenes that can be described by a single image, \textit{e.g.}, a small collection of objects on a tabletop or a sparse indoor furniture arrangement.

In this paper, we show that a complex scene containing hundreds of objects can be compositionally generated from a strong single-view, single-object generative prior. We take Pixal3D~\cite{li2026pixal3d} as a case study to demonstrate the feasibility and scalability of this paradigm. Our key idea is to retain its object-level generative prior while extending it with a multi-view conditioning pathway. For each object, we first construct an anchor-aligned canonical frame from its scene observations, then lift per-view DINOv3~\cite{simeoni2025dinov3} features into the corresponding canonical voxel volume. Features from different views are fused with an IBR-style aggregator~\cite{wang2021ibrnet,schmid2026genrecon} that is permutation-invariant and supports a variable number of inputs. The aggregated 3D condition is injected into the frozen prior through zero-initialized projection layers, while low-rank adapters (LoRA)~\cite{hu2022lora} adapt the pretrained network to exploit the additional multi-view evidence. As a result, unobserved regions can be plausibly generated rather than left incomplete, while the recovered geometry remains grounded in the available multi-view evidence. Each object is generated as an individual mesh in its anchor-aligned canonical frame and then placed into the scene through its canonical-to-world transformation, without cross-object fusion or joint shape optimization. Importantly, this formulation requires no scene-level training. The prior is finetuned entirely on individual objects in canonical space, yet generalizes at test time to large scenes containing hundreds of densely occluded objects. We further introduce a conditioning-view augmentation curriculum to improve robustness to the partial and degraded observations commonly encountered in cluttered scenes.

To enable meaningful evaluation on genuinely cluttered scenes, we introduce \emph{UE-MeshyScene}, a benchmark of large-scale indoor and outdoor environments rendered in Unreal Engine. Each scene contains up to several hundred assets arranged in natural configurations with substantial and complex mutual occlusion, together with annotated camera parameters, 3D bounding boxes, and ground-truth meshes for individual objects. UE-MeshyScene provides both the realism and scale of complex scenes and the exact geometric ground truth that is difficult to obtain from real-world captures. We evaluate our method on controlled single-object stress tests using Toys4k~\cite{stojanov2021using}, controlled multi-object scenes including the real-captured HouseCat6D~\cite{jung2024housecat6d} and synthetic Toys4k~\cite{stojanov2021using}-Scene datasets, as well as our UE-MeshyScene benchmark. Across all settings, our method consistently outperforms prior approaches, with larger gains as scenes become increasingly crowded and occluded.

In summary, our contributions are:
\begin{itemize}
\item We propose \emph{WorldSculpt}, a framework for generating compositional 3D scenes containing hundreds of objects, each represented by an individual mesh. WorldSculpt maps multi-view scene observations into an anchor-aligned canonical frame, conditions a strong object-level 3D generative prior on the aligned observations, and places the generated meshes into a shared world frame through canonical-to-world transformations.
\item We show that a generative prior finetuned entirely on individual objects in canonical space can generalize to large-scale scenes containing hundreds of densely occluded objects, without any scene-level training, through multi-view conditioning and a targeted augmentation curriculum.
\item We introduce \emph{UE-MeshyScene}, a photorealistic benchmark of densely cluttered scenes containing hundreds of objects, with exact per-object annotations and ground-truth meshes for evaluating compositional 3D generation under complex occlusion.
\end{itemize}

\section{Related Work}
\label{sec:related}

\paragraph{Image and multi-view to 3D generation.} Feed-forward 3D reconstruction methods directly infer geometry from one or a small number of input images. LRM~\cite{hong2024lrm} and subsequent mesh- and Gaussian-based variants, including MeshLRM~\cite{wei2024meshlrm}, InstantMesh~\cite{xu2024instantmesh}, GRM~\cite{xu2024grm}, LGM~\cite{tang2024lgm}, TripoSR~\cite{tochilkin2024triposr}, and SF3D~\cite{boss2025sf3d}, use large transformer-based architectures to predict meshes or Gaussian representations in a single forward pass. Another family of methods transfers 2D diffusion priors to 3D generation by first synthesizing novel views. Methods such as Zero-1-to-3~\cite{liu2023zero}, SyncDreamer~\cite{liu2024syncdreamer}, Wonder3D~\cite{long2024wonder3d}, CRM~\cite{wang2024crm}, and One-2-3-45~\cite{liu2023one} generate consistent multi-view observations that are subsequently fused into a 3D shape. Native-3D generative models instead learn distributions directly in 3D representations, including point clouds in Point-E~\cite{nichol2022point}, implicit functions in Shap-E~\cite{jun2023shap}, aligned shape-image-text latent spaces in Michelangelo~\cite{zhao2023michelangelo}, and more recent high-resolution structured representations used by TRELLIS~\cite{xiang2025structured}, Hunyuan3D~\cite{zhao2025hunyuan3d}, and Pixal3D~\cite{li2026pixal3d}. Such models are commonly trained on large-scale 3D asset datasets such as Objaverse and Objaverse-XL~\cite{deitke2023objaverse,deitke2023objaversexl}. Complementary pose-free stereo approaches, including DUSt3R~\cite{wang2024dust3r} and MASt3R~\cite{leroy2024grounding}, jointly estimate scene geometry and camera poses and can provide the camera information required by our pipeline.

\paragraph{Amodal 3D reconstruction.} Amodal 3D reconstruction seeks to recover the complete geometry of an object from partial observations affected by occlusion. Early work studies this problem in cluttered robotic environments, where incomplete 3D observations are completed using physical constraints such as object stability and connectivity~\cite{agnew2021amodal}. Another direction performs completion first in image space, as in pix2gestalt~\cite{ozguroglu2024pix2gestalt}, and then reconstructs the completed image with an image-to-3D model. However, image-space completion alone does not explicitly enforce 3D consistency or consistency across multiple views. More recently, Amodal3R~\cite{wu2025amodal3r} adapts a native-3D generative prior to directly recover complete object geometry from partially occluded images. Concurrent work, AmodalGen3D~\cite{zhou2025amodalgen3d}, extends this setting to amodal object generation from sparse, unposed views.

\paragraph{Object-level 3D datasets.} Evaluating image-to-3D methods requires datasets with ground-truth 3D geometry. Large synthetic or artist-created collections, including Objaverse and Objaverse-XL~\cite{deitke2023objaverse,deitke2023objaversexl} and Toys4k~\cite{stojanov2021using}, provide clean object meshes for both training and held-out evaluation. Google Scanned Objects~\cite{downs2022google} and ABO~\cite{collins2022abo} further provide scanned or catalog-based assets, although evaluation is typically performed on rendered images with clean backgrounds. Datasets such as CO3D~\cite{reizenstein2021common} and MVImgNet~\cite{yu2023mvimgnet} contain real captures with natural backgrounds and lighting, but provide SfM point clouds rather than ground-truth meshes. Datasets that pair real in-the-wild photographs with accurate scanned mesh geometry remain comparatively scarce.

\paragraph{Scene-level and compositional generation.} Multi-object scene generation has been studied from both single images and real-world captures. Holistic image-based approaches, including InstPIFu~\cite{liu2022towards}, PartCrafter~\cite{lin2026partcrafter}, MIDI~\cite{huang2025midi}, SAM3D~\cite{chen2026sam}, SceneGen~\cite{meng2026scenegen}, SceneMaker~\cite{shi2026scenemaker}, and RecGen~\cite{zadaianchuk2026recgen}, jointly recover object geometry and scene layout, and are commonly evaluated on synthetic datasets such as 3D-FRONT/3D-FUTURE~\cite{fu20213d,fu20213dfuture} and Hypersim~\cite{roberts2021hypersim}. Real-world datasets, in contrast, typically provide either fused but incomplete scene-level meshes, as in ScanNet~\cite{dai2017scannet}, ScanNet++~\cite{yeshwanth2023scannetpp}, Matterport3D~\cite{chang2017matterport3d}, MultiScan~\cite{mao2022multiscan}, and Replica~\cite{straub2019replica}, or object-level annotations in the form of oriented bounding boxes, as in ARKitScenes~\cite{baruch2021arkitscenes}, SUN RGB-D~\cite{song2015sun}, and Objectron~\cite{ahmadyan2021objectron}. Other datasets provide aligned CAD models as object proxies, including Scan2CAD~\cite{avetisyan2019scan2cad}, ROCA~\cite{gumeli2022roca}, and CAD-Estate~\cite{maninis2023cad}. Accurate and complete per-object mesh ground truth for real, cluttered scenes, however, remains difficult to obtain.

\paragraph{Structure and motion from multiple images.} Reconstructing objects from multiple images typically requires estimates of camera motion, coarse 3D structure, and object localization, for which a broad range of existing methods can be used. Camera poses are traditionally recovered with structure-from-motion~\cite{schoenberger2016sfm,schoenberger2016mvs,pan2024glomap} or deep visual SLAM~\cite{teed2021droid,teed2023deep}, while more recent systems target casual, low-parallax, and dynamic video specifically~\cite{li2025megasam,huang2025vipe}. Dense scene structure can be obtained from monocular geometry estimators, which have evolved from affine-invariant relative depth~\cite{ranftl2020towards,yang2024depth,yang2024depthv2,lin2025depth} to metric depth and camera-aware point maps jointly estimated with intrinsics~\cite{yin2023metric3d,piccinelli2024unidepth,wang2025moge}, as well as temporally consistent depth for long videos~\cite{hu2025depthcrafter,chen2025video}. More recently, feed-forward pointmap models have begun to unify camera estimation and geometry reconstruction by directly predicting both from unposed image sets or video streams~\cite{wang2024dust3r,leroy2024grounding,murai2025mast3r,zhang2025monst3r,yang2025fast3r,wang2025continuous,wang2025vggt,keetha2026mapanything,wang2025pi,lin2025depth}. Object-level processing can be handled separately using promptable or open-world video segmentation~\cite{ravi2025sam,carion2025sam,cheng2023tracking,cheng2024putting,cheng2022xmem}, optionally initialized by open-vocabulary detectors~\cite{liu2024grounding,ren2024grounded,wu2024general,cheng2024yolo}, to obtain temporally consistent object masks. These masks can then be combined with 3D detection or object-level SLAM methods~\cite{brazil2023omni3d,rukhovich2022imvoxelnet,zhang2025detect,lazarow2025cubify,detone2026boxer,yang2019cubeslam,li2021odam,wen2023bundlesdf,lemeshko2026zoo3d}, or with classical silhouette-based geometry when camera poses are known~\cite{laurentini1994visual,kutulakos2000theory}, to recover object locations in 3D. Taken together, these components can convert raw monocular footage into camera trajectories, dense scene geometry, object mask tracks, and object-level 3D localizations.

\section{Method}
\label{sec:method}

\paragraph{Problem setup.}
Given a set of $N$ posed images $\{I_n\}_{n=1}^{N}$, we assume known camera intrinsics $\{K_n\}_{n=1}^{N}$ and camera-to-world extrinsics $\{T_n^{\mathrm{cw}}\}_{n=1}^{N}$. For each object $k$, we assume access to a per-view instance mask $S_{kn}$ whenever the object is visible~\cite{carion2025sam,ravi2025sam,cheng2023tracking,cheng2024putting,cheng2022xmem,wu2024general,liu2024grounding,ren2024grounded,cheng2024yolo}, together with a coarse world-space localization box $B_k^{\mathrm{loc}}$~\cite{brazil2023omni3d,rukhovich2022imvoxelnet,zhang2025detect,lazarow2025cubify,detone2026boxer,lemeshko2026zoo3d,yang2019cubeslam,li2021odam,wen2023bundlesdf}. Recovering these masks, camera parameters, and coarse object localizations is well studied and lies outside the scope of this work.

Our goal is to produce a compositional scene representation
\begin{equation}
    \mathcal{M}
    =
    \left\{
    \left(
    \mathcal{M}_k^{\mathrm{c}},
    T_k^{\mathrm{ow}}
    \right)
    \right\}_{k=1}^{K},
\end{equation}
where $\mathcal{M}_k^{\mathrm{c}}$ is an individual mesh generated in the canonical frame of object $k$, and $T_k^{\mathrm{ow}}$ maps that canonical frame into the world coordinate system. The corresponding world-space mesh is
\begin{equation}
    \mathcal{M}_k^{\mathrm{w}}
    =
    T_k^{\mathrm{ow}}
    \left(
    \mathcal{M}_k^{\mathrm{c}}
    \right).
\end{equation}
The scene remains a collection of individually addressable object meshes rather than a single fused representation, making it directly suitable for downstream rendering, simulation, and content-authoring pipelines.

\begin{figure}[!t]
  \centering
  \includegraphics[width=\linewidth]{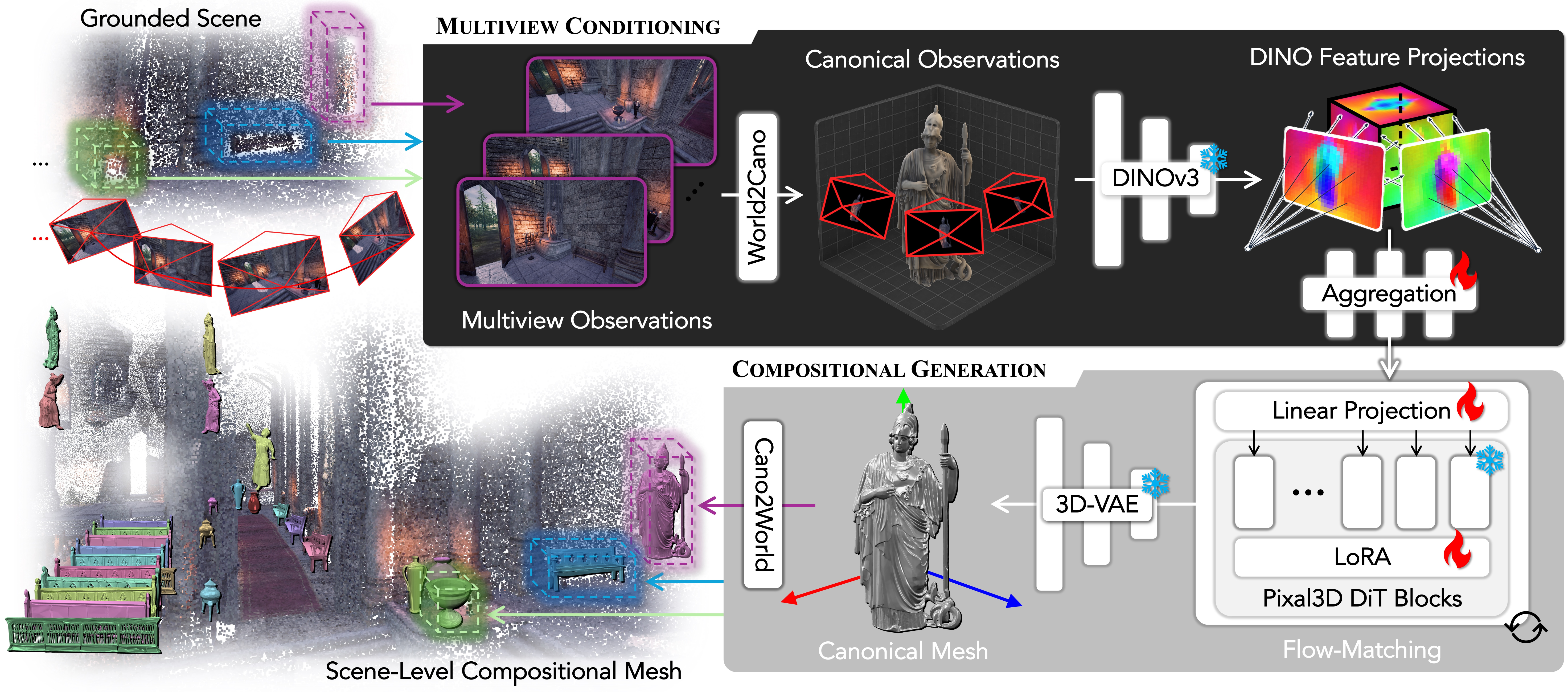}\vspace{5pt}
  \caption{\textbf{Overview of WorldSculpt.}
  Given posed scene images, instance masks, and coarse 3D object localizations, the anchor-aligned canonicalization stage constructs a virtual canonical cube for each object and derives crop-aware projections from its input views. Per-view DINOv3 features are lifted into the canonical voxel volume and fused with a permutation-invariant aggregator. The resulting 3D condition, together with global tokens from the anchor view, drives the two geometry stages of a Pixal3D prior through zero-initialized injection layers and LoRA adaptation. The generated canonical meshes are finally placed into the scene using their canonical-to-world transformations.}
  \label{fig:pipeline}
\end{figure}

Our method consists of three main steps. We first construct an anchor-aligned virtual canonical frame for each object and map its scene observations into that frame. We then generate an individual object mesh with a multi-view conditioned 3D generative prior. Finally, the generated mesh is transformed back into the world frame using the same canonical-to-world transformation. Figure~\ref{fig:pipeline} provides an overview.

\subsection{Anchor-aligned object canonicalization}
\label{sec:method:canonical}

\paragraph{Anchor view and canonical cube.}
The supplied coarse localization box $B_k^{\mathrm{loc}}$ is not directly used as the generation volume. Its side lengths are generally unequal, while the generative model operates in a normalized cubic domain. Its orientation is also determined by the localization procedure and need not agree with the camera-relative canonical orientation expected by the pretrained object prior.

For each object, we therefore construct an anchor-aligned virtual canonical cube. Let $\mathcal{I}_k$ denote the set of views in which object $k$ is observed. We select one view $a_k \in \mathcal{I}_k$ as the \emph{anchor}. At inference time, the anchor is chosen as the view in which the object is most fully observed. During training, it is sampled randomly from the available views. The anchor determines the orientation of the virtual canonical frame, with the anchor camera observing the object from the canonical front-view direction.

We use the normalized cube
\begin{equation}
    \mathcal{V}
    =
    \left[-\frac{1}{2},\frac{1}{2}\right]^3
\end{equation}
as the canonical spatial domain. Let $c_k$ be the center of $B_k^{\mathrm{loc}}$, and let $s_k$ initially be its largest side length. The rotation $R_k \in \mathrm{SO}(3)$ is induced by the anchor camera orientation so that the anchor view is mapped to the canonical viewing convention. The canonical-to-world transformation is then
\begin{equation}
    T_k^{\mathrm{ow}}
    =
    \begin{bmatrix}
        s_k R_k & c_k \\
        0 & 1
    \end{bmatrix},
    \label{eq:canonical_transform}
\end{equation}
which maps a canonical point $x \in \mathcal{V}$ to
\begin{equation}
    x_k^{\mathrm{w}}
    =
    s_k R_k x + c_k.
\end{equation}
Since $s_k$ is isotropic, $T_k^{\mathrm{ow}}$ is a similarity transformation that preserves the proportions of the generated object.

Localization boxes may be inaccurate (\ie, slightly loose or tight). We therefore allow $s_k$ to increase while keeping $c_k$ and $R_k$ fixed until the projection of the resulting cube covers the object masks in all selected views. This adjustment prevents the object from being clipped by the per-view crops while preserving a single consistent canonical frame across observations.

\paragraph{Canonical observations and crop-aware projection.}
For each selected view $n \in \mathcal{I}_k$, we project the anchor-aligned cube into the image, crop the image to its projected extent, mask out the background using $S_{kn}$, and resize the crop to the input resolution of the generative model. We denote the resulting object-centric image by $\bar I_{kn}$.

Cropping and resizing change the image coordinate system. Let $A_{kn}$ denote the corresponding transformation from the original image coordinates to the coordinates of the resized crop. The adjusted camera intrinsics are
\begin{equation}
    \bar K_{kn}
    =
    A_{kn} K_n.
\end{equation}
For a canonical voxel $x$, with homogeneous coordinate
$\tilde{x} = [x^\top,1]^\top$, its location in the cropped observation is
\begin{equation}
    \pi_{kn}(x)
    =
    \Pi
    \left(
        \bar K_{kn}
        \begin{bmatrix}
            I_3 & 0
        \end{bmatrix}
        \left(T_n^{\mathrm{cw}}\right)^{-1}
        T_k^{\mathrm{ow}}
        \tilde{x}
    \right),
    \label{eq:canonical_projection}
\end{equation}
where $\Pi([u,v,d]^\top)=(u/d,v/d)^\top$ denotes perspective division. Equation~\eqref{eq:canonical_projection} first maps the voxel from the object canonical frame into the world frame, then into the camera frame, and finally into the resized crop. It therefore accounts for the object's position, orientation, scale, and off-center image location without assuming that the object is centered in any input view.

\subsection{Multi-view conditioned object generation}
\label{sec:method:mv}

\paragraph{Generative prior.}
We instantiate the per-object generator $\Phi$ from Pixal3D~\cite{li2026pixal3d}, a native-3D generative model built on the TRELLIS.2 structured-latent backbone~\cite{xiang2025trellis2}. Pixal3D generates an object through a cascade of flow-matching stages~\cite{lipman2022flow}. We use its first two stages, which determine object geometry. The \emph{sparse structure} stage predicts coarse occupancy on a $64^3$ grid, and the \emph{shape} stage generates a high-resolution sparse structured latent over the occupied voxels. The resulting latents are decoded into a mesh. We focus on geometry and leave the subsequent texture and material stage to future work.

The original Pixal3D model consumes a single pose-free RGB image through cross-attention. Its output orientation, position, and scale are therefore implicit in the single-image training convention rather than explicitly tied to camera geometry. Our setting instead provides a coherent set of posed observations that must jointly constrain the generated object in a known frame. We retain the pretrained object prior and introduce a multi-view conditioning pathway that grounds its generation in the anchor-aligned canonical volume.

\paragraph{Multi-view feature lifting.}
Each canonical observation $\bar I_{kn}$ is independently encoded with DINOv3~\cite{simeoni2025dinov3}, producing a dense feature map
\begin{equation}
    F_{kn}
    =
    E_{\mathrm{DINO}}
    \left(
        \bar I_{kn}
    \right).
\end{equation}
For each voxel $x$ in the canonical volume, the feature from view $n$ is sampled at its projected image location:
\begin{equation}
    g_n(x)
    =
    F_{kn}
    \left(
        \pi_{kn}(x)
    \right).
    \label{eq:feature_lifting}
\end{equation}
This operation lifts each 2D feature map into a 3D feature grid whose spatial coordinates are shared across all views. Observations of the same canonical location are therefore aligned before cross-view aggregation. 

\paragraph{Permutation-invariant view aggregation.}
Let $\mathcal{J}_k \subseteq \mathcal{I}_k$ be the selected conditioning views for object $k$, and let $L_k = |\mathcal{J}_k|$. We aggregate the lifted features at each voxel using an IBRNet-style module~\cite{schmid2026genrecon,wang2021ibrnet}. The module accepts a variable number of views and does not depend on their ordering.

For each canonical voxel $x$, we first compute the cross-view mean and variance:
\begin{equation}
    \mu(x)
    =
    \frac{1}{L_k}
    \sum_{n \in \mathcal{J}_k}
    g_n(x),
\end{equation}
\begin{equation}
    \sigma^2(x)
    =
    \frac{1}{L_k}
    \sum_{n \in \mathcal{J}_k}
    g_n(x) \odot g_n(x)
    -
    \mu(x) \odot \mu(x),
\end{equation}
where $\odot$ denotes element-wise multiplication. Two lightweight MLPs then refine each feature and predict its aggregation logit:
\begin{equation}
    g_n'(x)
    =
    \mathrm{MLP}_{\mathrm{feat}}
    \left(
        \left[
        g_n(x),
        \mu(x),
        \sigma^2(x)
        \right]
    \right),
\end{equation}
\begin{equation}
    w_n(x)
    =
    \mathrm{MLP}_{\mathrm{weight}}
    \left(
        \left[
        g_n(x),
        \mu(x),
        \sigma^2(x)
        \right]
    \right).
\end{equation}
The final voxel feature is the cross-view mean plus a softmax-weighted residual:
\begin{equation}
    g_{\mathrm{out}}(x)
    =
    \mu(x)
    +
    \sum_{n \in \mathcal{J}_k}
    \alpha_n(x) g_n'(x),
    \qquad
    \alpha_n(x)
    =
    \frac{\exp(w_n(x))}
    {\sum_{m \in \mathcal{J}_k}\exp(w_m(x))}.
    \label{eq:ibr}
\end{equation}
The final layer of $\mathrm{MLP}_{\mathrm{feat}}$ is initialized to zero. The aggregator therefore begins as an exact cross-view mean and gradually learns view-dependent residual corrections during training. Applying this operation to every voxel produces the aggregated 3D conditioning grid $G_k$.

All selected views contribute to $G_k$, while the global image tokens used by the original Pixal3D cross-attention interface are extracted only from the anchor observation $\bar I_{k a_k}$. The anchor tokens preserve the single-image conditioning convention of the pretrained model, and the voxel-aligned grid introduces additional evidence from the remaining views.

\paragraph{Conditioning injection.}
We inject $G_k$ into both Pixal3D geometry stages. At each transformer block, the aggregated condition is aligned with the current 3D token locations, projected to the block feature dimension, and added to the corresponding block features. We keep the original Pixal3D parameters frozen and adapt its attention and per-block projection layers using low-rank adapters (LoRA)~\cite{hu2022lora}. The multi-view aggregator and the zero-initialized conditioning projections are trained in full. This design retains the completion ability of the pretrained object prior while allowing the denoiser to use spatially aligned evidence from multiple views.

\subsection{Training}
\label{sec:method:train}

\paragraph{Random-view training.}
The sparse-structure and shape stages are trained independently using their respective ground-truth latent targets. For each training object, we sample a variable number of conditioning views and randomly choose one of them as the anchor. The sampled anchor defines the camera-relative canonical orientation for that training example, and the conditioning views and target latents are expressed in the same anchor-aligned frame. Varying the anchor and the number of observations exposes the model to different camera configurations while the aggregation architecture remains permutation-invariant.

Only the multi-view aggregator, the conditioning injection layers, and the LoRA parameters are optimized. All remaining parameters of the pretrained Pixal3D generator stay frozen.

\paragraph{Conditioning-view augmentation.}
The clean object renders used for training differ substantially from the partial and degraded observations found in cluttered scenes. We reduce this gap with a conditioning-view augmentation curriculum. The augmentations are applied only to the input observations and never to the supervision target.

We independently sample four types of degradation. First, we simulate occlusion using either random 2D masks or 3D-consistent occluders placed between the camera and the target object. The latter produce occlusion patterns that move coherently across views. Second, we perturb the camera poses of the non-anchor observations while keeping the anchor pose unchanged. Third, we introduce mask errors by perturbing and degrading the segmentation boundaries. Fourth, we downsample the observations to simulate objects that occupy only a small image region. The strength of each degradation is gradually increased from zero to its maximum value during training, allowing the model to first learn from clean observations before encountering more difficult inputs.

\subsection{Inference and scene composition}
\label{sec:method:test}

At inference time, we process each object independently. We first select the anchor as the view in which the object is most fully observed and construct the anchor-aligned canonical cube described in Section~\ref{sec:method:canonical}. If the initial cube does not cover the object masks in all candidate views, its side length is increased while its center and orientation remain fixed. We then select a bounded number of conditioning views, prioritizing observations with larger visible object regions, and produce the corresponding masked crops and crop-aware projection functions.

Each object is generated by running the two Pixal3D geometry stages with the multi-view conditioning described in Section~\ref{sec:method:mv}. The sparse-structure stage predicts the coarse occupied voxels, which define the sparse support used by the shape stage. The second-stage latent is then decoded into an individual canonical-space mesh $\mathcal{M}_k^{\mathrm{c}}$.

The mesh is placed into the scene using the canonical-to-world transformation from Eq.~\eqref{eq:canonical_transform}:
\begin{equation}
    \mathcal{M}_k^{\mathrm{w}}
    =
    T_k^{\mathrm{ow}}
    \left(
        \mathcal{M}_k^{\mathrm{c}}
    \right).
\end{equation}
The final compositional scene is the collection
\begin{equation}
    \mathcal{M}^{\mathrm{w}}
    =
    \left\{
        \mathcal{M}_k^{\mathrm{w}}
    \right\}_{k=1}^{K}.
\end{equation}

\subsection{Implementation details}
\label{sec:method:implementation}

We initialize the two Pixal3D geometry stages from their released checkpoints and finetune them independently on the TexVerse dataset. Conditioning images are resized to $512$px for the sparse-structure stage and $1024$px for the shape stage. The LoRA adapters use rank $r{=}32$ and scaling factor $\alpha{=}32$ and are applied to the DiT attention and per-block projection layers. The multi-view aggregator and the zero-initialized conditioning projections are optimized without low-rank parameterization.

Each stage is trained for $15$k iterations in \texttt{bf16} using AdamW with a learning rate of $10^{-4}$, betas $(0.9,0.95)$, no weight decay, and an exponential moving average of $0.9999$. We use a batch size of $48$ for the sparse-structure stage and $12$ for the shape stage. At each training iteration, we randomly sample $1$--$20$ views to form the multi-view conditioning set. To limit memory usage, the shape stage applies gradient checkpointing to half of its transformer blocks and caps each conditioning view at $32{,}768$ tokens. The augmentation curriculum reaches its maximum strength over the first $3$k training iterations.
\section{Experiments}
\label{sec:experiments}

\subsection{Datasets and Metrics}

We evaluate our method under three settings of increasing difficulty.
(i) We first evaluate canonical-space generation of individual objects on Toys4k~\cite{stojanov2021using} (Sec.~\ref{sec:exp:canon}), isolating the ability of the model to recover object geometry from partial multi-view observations.
(ii) We then evaluate compositional scene generation on synthetic Toys4k-Scene and the real-world HouseCat6D~\cite{jung2024housecat6d} dataset (Sec.~\ref{sec:exp:multi}).
(iii) Finally, we evaluate large-scale compositional generation on UE-MeshyScene, where scenes contain hundreds of objects under severe mutual occlusion (Sec.~\ref{sec:exp:ue}).

\paragraph{Datasets.}
\emph{Toys4k}~\cite{stojanov2021using} consists of clean, isolated object models and is used for controlled single-object evaluation. We vary both the number of input views ($1$--$16$) and the per-view occlusion fraction ($0$--$75\%$), allowing us to measure how generation quality changes as observations become fewer and increasingly incomplete, without introducing scene-level factors.

\emph{Toys4k-Scene} is a synthetic benchmark constructed by placing multiple Toys4k objects into cluttered layouts and rendering short orbiting sequences. It preserves the clean ground-truth geometry of Toys4k while introducing substantial inter-object occlusion and challenging multi-object configurations. Many objects contain thin or intricate structures and are densely arranged, leaving only partial observations from typical viewpoints. This setting is therefore particularly challenging for methods that rely on a single image.

\emph{HouseCat6D}~\cite{jung2024housecat6d} provides a complementary real-world evaluation setting. It contains real captures of tabletop scenes together with scanned ground-truth meshes for individual objects. Compared with Toys4k-Scene, its layouts are relatively sparse, the objects are mostly geometrically simple household items, and inter-object occlusion is generally mild. It therefore allows us to evaluate how well a model trained on synthetic renders transfers to real imagery.

\emph{UE-MeshyScene} is our large-scale benchmark for evaluating compositional generation in densely cluttered scenes. We describe its construction and statistics in detail below.

\begin{table}[!t]\centering
\setlength{\abovecaptionskip}{2mm}
\caption{\textbf{Per-scene statistics of UE-MeshyScene.} Every scene is rendered
in Unreal Engine~5.8 at $2560\times1440$; ``\#Views'' denotes the number of
rendered frames.}
\label{tab:ue_config}
\setlength{\tabcolsep}{8pt}
\resizebox{\linewidth}{!}{%
\begin{tabular}{lccccccc}
\toprule
 & \textit{Hangar} & \textit{Abandoned City} & \textit{Cathedral} & \textit{Office} & \textit{Japanese School} & \textit{Desert Town} & Total \\
\midrule
\#Objects & 171 & 93 & 145 & 678 & 511 & 701 & 2{,}299 \\
\#Views   & 463 & 265 & 956 & 1268 & 1758 & 1254 & 5{,}964 \\
\bottomrule
\end{tabular}
}
\end{table}

\begin{figure}[!t]
\centering
\includegraphics[width=\linewidth]{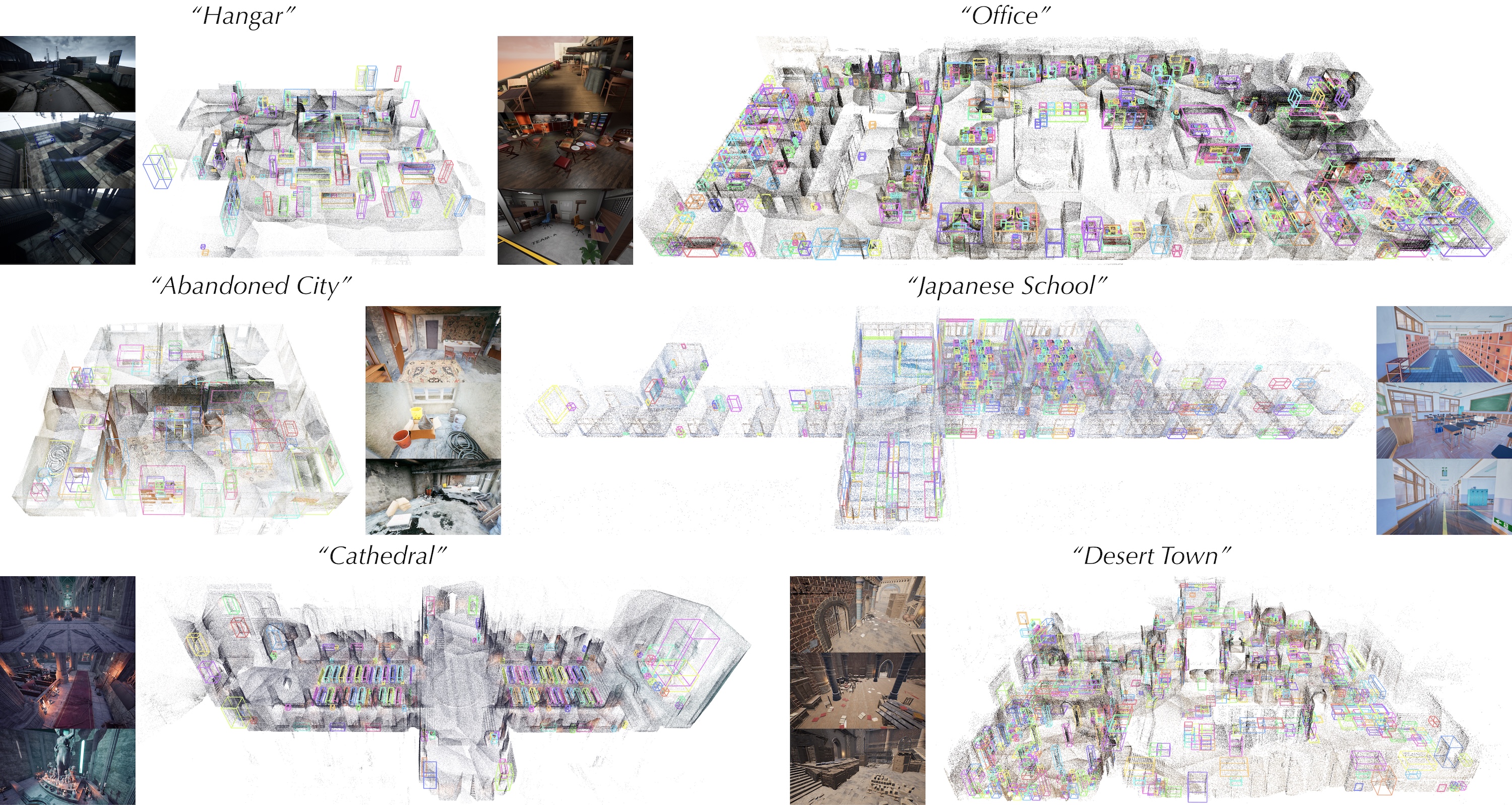}\vspace{5pt}
\caption{\textbf{Visualization of the UE-MeshyScene dataset.} Our benchmark contains six photorealistic environments spanning relatively orderly to extremely cluttered layouts, with up to $701$ objects per scene. Each environment is captured along an orbiting camera trajectory and provides exact per-object ground truth, enabling world-frame evaluation under complex mutual occlusion.}
\label{fig:datasample}
\end{figure}

\paragraph{UE-MeshyScene.}
Existing scene benchmarks remain limited in either scale, occlusion, or the availability of individual per-object ground truth. HouseCat6D contains relatively sparse layouts with mild inter-object occlusion, while Toys4k-Scene introduces substantially stronger occlusion but still contains only a modest number of objects. Realistic environments are considerably more challenging, as they may contain dense clutter and large numbers of mutually occluding objects. To evaluate this setting, we introduce UE-MeshyScene, a collection of six large-scale scenes authored and rendered in Unreal Engine. The scenes contain between $93$ and $701$ objects arranged naturally across shelves, desks, floors, and other cluttered regions, producing complex patterns of mutual occlusion. Since every asset is a known 3D model placed with a known transformation, UE-MeshyScene provides complete per-object mesh ground truth, exact camera poses, and 3D bounding boxes. This enables direct evaluation in a common world coordinate frame while retaining the complexity of densely cluttered environments.

UE-MeshyScene comprises six environments authored and rendered in Unreal Engine~5.8: an aircraft hangar, an abandoned-city interior, an old cathedral, an architectural-visualization office, a Japanese school, and a desert town. Together, they span a range of layouts from relatively orderly to extremely cluttered. Each scene is rendered along a smooth camera trajectory at a resolution of $2560\times1440$, producing $265$--$1758$ views per scene and $5{,}964$ views in total. The scenes contain between $93$ and $701$ objects each, with $2{,}299$ objects overall. Table~\ref{tab:ue_config} summarizes the statistics of each scene, and Figure~\ref{fig:datasample} shows representative examples. For every frame, the dataset provides an RGB image, camera pose, per-object instance masks, per-object 3D bounding boxes, and a metric depth map. Although depth is not used by our method, we release the full set of annotations to support related tasks.

\begin{table*}[t]\centering
\setlength{\abovecaptionskip}{2mm}
\caption{Canonical-space generation on Toys4k~\cite{stojanov2021using} under varying
numbers of input views (\#V) and per-view occlusion levels. We report
CD-$\ell_2$ ($\times10^{-3}$, $\downarrow$), CD-$\ell_1$ ($\times10^{-2}$, $\downarrow$),
EMD ($\times10^{-2}$, $\downarrow$), and F-Score@$\{.02,.05\}$ ($\uparrow$).
The best result in each column is shown in \textbf{bold}.}
\label{tab:main}
\setlength{\tabcolsep}{3pt}
\resizebox{\linewidth}{!}{%
\begin{tabular}{l c *{20}{c}}
\toprule
 & & \multicolumn{5}{c}{Occl.\ $0\%$} & \multicolumn{5}{c}{Occl.\ $25\%$}
   & \multicolumn{5}{c}{Occl.\ $50\%$} & \multicolumn{5}{c}{Occl.\ $75\%$} \\
\cmidrule(lr){3-7}\cmidrule(lr){8-12}\cmidrule(lr){13-17}\cmidrule(lr){18-22}
Method & \#V
 & CD-$\ell_2$ & CD-$\ell_1$ & EMD & F@.02 & F@.05
 & CD-$\ell_2$ & CD-$\ell_1$ & EMD & F@.02 & F@.05
 & CD-$\ell_2$ & CD-$\ell_1$ & EMD & F@.02 & F@.05
 & CD-$\ell_2$ & CD-$\ell_1$ & EMD & F@.02 & F@.05 \\
\midrule
TRELLIS.2~\cite{xiang2025trellis2} & 1
 & 7.55 & 6.35 & 10.7 & 0.587 & 0.839 & 14.19 & 8.86 & 15.1 & 0.480 & 0.739
 & 27.09 & 13.17 & 20.4 & 0.344 & 0.587 & 56.71 & 20.63 & 28.5 & 0.189 & 0.367 \\
Pixal3D~\cite{li2026pixal3d} & 1
 & 2.17 & 3.74 & 7.8 & 0.749 & 0.934
 & 7.68 & 6.80 & 15.2 & 0.556 & 0.813
 & 22.81 & 11.69 & 21.6 & 0.405 & 0.651
 & 63.95 & 21.51 & 31.8 & 0.213 & 0.401 \\
\midrule
TRELLIS~\cite{xiang2025structured} & 1
 & 9.39 & 6.88 & 11.1 & 0.556 & 0.824
 & 14.18 & 8.69 & 13.3 & 0.466 & 0.753
 & 25.02 & 11.89 & 16.9 & 0.375 & 0.645
 & 44.46 & 17.16 & 22.5 & 0.256 & 0.487 \\
 & 2
 & 8.29 & 6.50 & 10.7 & 0.574 & 0.837
 & 14.87 & 8.62 & 13.0 & 0.474 & 0.758
 & 21.31 & 11.34 & 16.3 & 0.368 & 0.649
 & 39.15 & 16.58 & 22.1 & 0.244 & 0.481 \\
 & 4
 & 7.74 & 6.29 & 10.5 & 0.583 & 0.845
 & 13.07 & 8.16 & 12.6 & 0.492 & 0.773
 & 19.58 & 10.65 & 15.5 & 0.386 & 0.672
 & 38.49 & 16.51 & 21.8 & 0.235 & 0.471 \\
 & 8
 & 8.91 & 6.51 & 10.7 & 0.579 & 0.839
 & 11.53 & 7.73 & 12.2 & 0.508 & 0.789
 & 18.05 & 10.25 & 15.1 & 0.399 & 0.683
 & 33.81 & 15.68 & 21.4 & 0.241 & 0.480 \\
 & 16
 & 9.46 & 6.65 & 10.8 & 0.577 & 0.835
 & 11.59 & 7.73 & 12.0 & 0.510 & 0.787
 & 18.51 & 10.29 & 15.1 & 0.401 & 0.687
 & 34.39 & 15.71 & 21.2 & 0.249 & 0.484 \\
\midrule
Ours & 1
 & 2.59 & 3.85 & 7.8 & 0.739 & 0.932
 & 3.70 & 4.73 & 8.7 & 0.669 & 0.896
 & 8.90 & 6.84 & 11.2 & 0.554 & 0.818
 & 25.45 & 12.30 & 16.8 & 0.370 & 0.631 \\
 & 2
 & 1.54 & 3.23 & 7.2 & 0.790 & 0.954
 & 2.57 & 3.87 & 7.9 & 0.734 & 0.931
 & 4.01 & 4.85 & 8.8 & 0.651 & 0.893
 & 10.03 & 7.96 & 12.1 & 0.494 & 0.762 \\
 & 4
 & 1.35 & 3.02 & 7.0 & 0.813 & 0.964
 & 1.62 & 3.35 & 7.3 & 0.778 & 0.950
 & 2.56 & 4.01 & 7.8 & 0.719 & 0.925
 & 5.67 & 5.86 & 10.1 & 0.590 & 0.849 \\
 & 8
 & 1.23 & 2.94 & 6.9 & 0.822 & 0.965
 & 1.50 & 3.20 & 7.2 & 0.796 & 0.956
 & 1.91 & 3.57 & 7.5 & 0.755 & 0.943
 & 3.52 & 4.65 & 8.6 & 0.663 & 0.902 \\
 & 16
 & \textbf{1.19} & \textbf{2.89} & \textbf{6.9} & \textbf{0.827} & \textbf{0.967}
 & \textbf{1.29} & \textbf{3.01} & \textbf{7.0} & \textbf{0.813} & \textbf{0.964}
 & \textbf{1.44} & \textbf{3.23} & \textbf{7.2} & \textbf{0.788} & \textbf{0.955}
 & \textbf{2.25} & \textbf{3.97} & \textbf{8.0} & \textbf{0.713} & \textbf{0.929} \\
\bottomrule
\end{tabular}
}
\end{table*}

\begin{table}[t]\centering
\begin{minipage}[t]{0.52\linewidth}\centering
\setlength{\abovecaptionskip}{2mm}
\caption{Compositional scene generation on the real-world HouseCat6D~\cite{jung2024housecat6d} dataset and synthetic Toys4k~\cite{stojanov2021using}-Scene benchmark. We report CD-$\ell_2$ ($\times10^{-3}$, $\downarrow$), CD-$\ell_1$ ($\times10^{-2}$, $\downarrow$), EMD ($\times10^{-2}$, $\downarrow$), and F-Score ($\uparrow$). The best result in each column is shown in \textbf{bold}.}
\label{tab:scene}
\setlength{\tabcolsep}{1pt}
\resizebox{\linewidth}{!}{%
\begin{tabular}{l *{8}{c}}
\toprule
Method & \multicolumn{4}{c}{HouseCat6D~\cite{jung2024housecat6d}}
   & \multicolumn{4}{c}{Toys4k-Scene~\cite{stojanov2021using}} \\
\cmidrule(lr){2-5}\cmidrule(lr){6-9}
 & CD-$\ell_2$$\downarrow$ & CD-$\ell_1$$\downarrow$ & EMD$\downarrow$ & F-Score$\uparrow$
 & CD-$\ell_2$$\downarrow$ & CD-$\ell_1$$\downarrow$ & EMD$\downarrow$ & F-Score$\uparrow$ \\
\midrule
SAM3D~\cite{chen2026sam}
 & 17.7 & 11.45 & 12.1 & 0.630
 & 52.7 & 17.45 & 17.5 & 0.457 \\
SceneGen~\cite{meng2026scenegen}
 & 31.4 & 13.08 & 15.9 & 0.626 & 25.3 & 11.79 & 15.6 & 0.664 \\
SceneMaker~\cite{shi2026scenemaker}
 & 16.3 & 12.98 & 12.3 & 0.519
 & 26.8 & 16.85 & 17.3 & 0.422 \\
RecGen~\cite{zadaianchuk2026recgen}
 & 7.22 & 6.98 & 8.6 & 0.805
 & 13.2 & 9.74 & 11.0 & 0.676 \\
\midrule
ShapeR~\cite{siddiqui2026shaperrobustconditional3d}
 & 1.26 & 1.88 & 4.8 & 0.973
 & 8.38 & 7.81 & 9.4 & 0.746 \\
Ours
 & \textbf{0.28} & \textbf{1.53} & \textbf{3.4} & \textbf{0.995}
 & \textbf{0.61} & \textbf{1.78} & \textbf{3.7} & \textbf{0.981} \\
\bottomrule
\end{tabular}
}
\end{minipage}\hfill
\begin{minipage}[t]{0.46\linewidth}\centering
\setlength{\abovecaptionskip}{2mm}
\caption{Compositional generation on six scenes of the \emph{UE-MeshyScene} dataset. Protocol and metrics as in Table~\ref{tab:scene} (F-Score at
$\tau{=}0.05$), additionally reporting per-instance medians
(CD$^{\mathrm{med}}$, same scale as the corresponding mean). Best per column
in \textbf{bold}.}
\label{tab:ue}
\setlength{\tabcolsep}{2pt}
\resizebox{\linewidth}{!}{%
\begin{tabular}{l *{6}{c}}
\toprule
Method & CD-$\ell_2$$\downarrow$ & CD-$\ell_2^{\mathrm{med}}$$\downarrow$
 & CD-$\ell_1$$\downarrow$ & CD-$\ell_1^{\mathrm{med}}$$\downarrow$
 & EMD$\downarrow$ & F-Score$\uparrow$ \\
\midrule
ShapeR~\cite{siddiqui2026shaperrobustconditional3d}
 & 7.42 & 2.47 & 6.37 & 5.07 & 8.8 & 0.813 \\
Ours
 & \textbf{2.48} & \textbf{0.25} & \textbf{2.66} & \textbf{1.69}
 & \textbf{5.0} & \textbf{0.951} \\
\bottomrule
\end{tabular}
}
\end{minipage}
\end{table}

\begin{figure}[!t]\centering
\includegraphics[width=\linewidth]{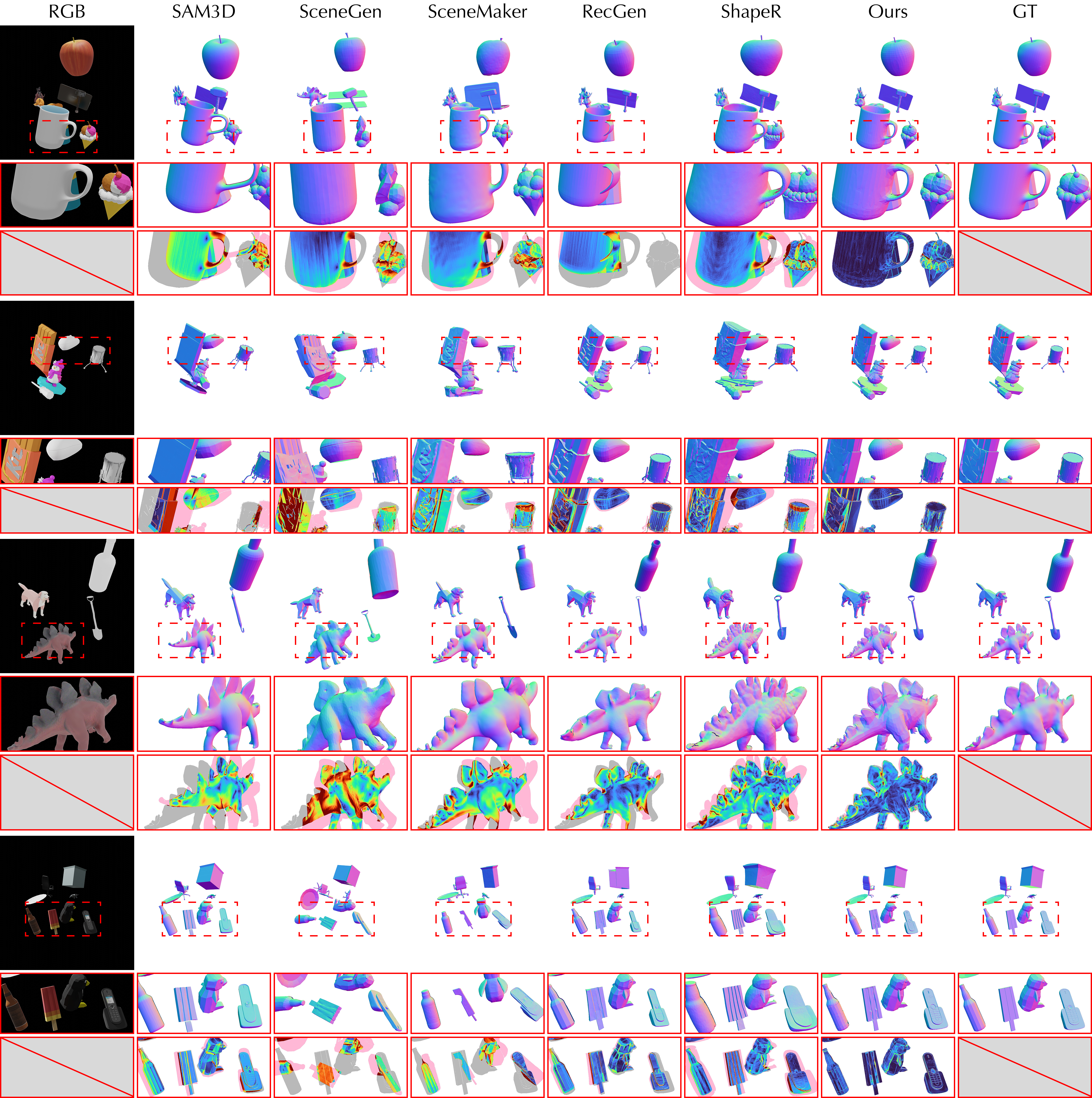}
\setlength{\abovecaptionskip}{-2mm}
\caption{\textbf{Qualitative comparison on Toys4k-Scene.}
Given the input observations, each method generates per-object meshes, whose surface normals are rendered in the shared world frame. Our method produces more detailed and accurately registered object geometry, closely matching the ground truth in the rightmost column. For each example, we enlarge the boxed region and visualize its pixelwise discrepancy with the ground truth in the row below. On overlapping surfaces, color denotes the angular error between surface normals, ranging from blue at \(0^\circ\) to red at \(90^\circ\). Non-overlapping regions are shown separately: \emph{pink} indicates spurious predicted surfaces, while \emph{gray} indicates missing ground-truth surfaces. These regions also contribute to the mask-IoU error.}
\label{fig:toys4k_scene}
\end{figure}

\begin{figure}[!t]\centering
\includegraphics[width=\linewidth]{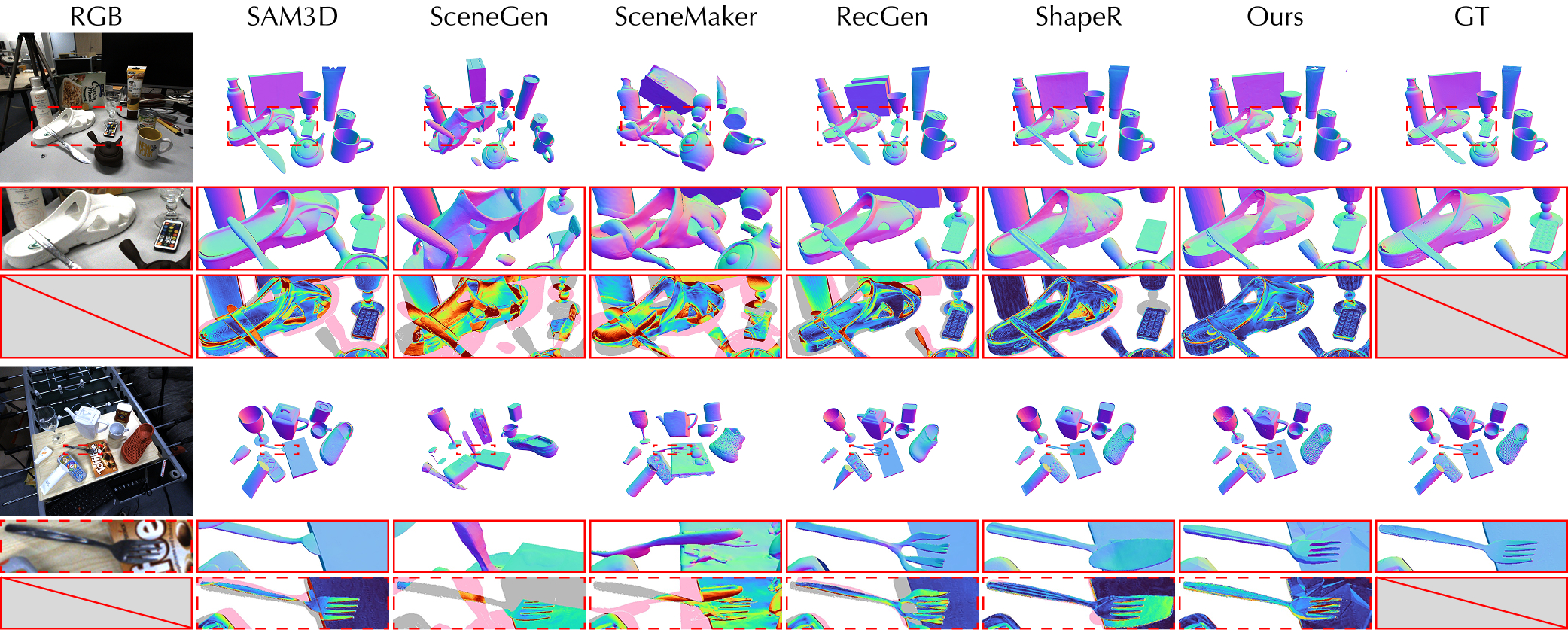}
\setlength{\abovecaptionskip}{-2mm}
\caption{\textbf{Qualitative comparison on HouseCat6D.}
Our method recovers detailed object geometry that is accurately placed in the world frame and closely matches the scanned ground truth shown in the rightmost column, while the baselines often exhibit incomplete or distorted shapes and inaccurate placement. As in Fig.~\ref{fig:toys4k_scene}, each example includes an enlarged region of interest and a pixelwise discrepancy map with respect to the ground truth in the row below. On overlapping surfaces, color indicates the angular error between surface normals, ranging from blue at $0^\circ$ to red at $90^\circ$. Non-overlapping regions are shown separately: \emph{pink} denotes spurious predicted surfaces, while \emph{gray} denotes missing ground-truth surfaces. Both contribute to the mask IoU error.}
\label{fig:housecat6d}
\end{figure}

\paragraph{Metrics.}
For each object, we decode the prediction into a mesh and compare it with the corresponding ground-truth geometry. We report Chamfer distance using both $\ell_2$ and $\ell_1$ distances (CD-$\ell_2$ and CD-$\ell_1$), Earth Mover's Distance (EMD), and F-Score.

For single-object evaluation in canonical space, both the prediction and ground truth are normalized with respect to the ground-truth unit sphere. Since different baselines adopt different canonical coordinate conventions, we perform per-instance ICP alignment before computing the metrics, making the comparison invariant to differences in canonical frame.

For compositional scene evaluation, each generated object is transformed from its canonical frame into the world frame using the canonical-to-world transformation. We then compare it directly with the corresponding ground-truth mesh in the shared world frame, without per-object ICP alignment. Distance-based metrics are normalized by the diagonal length of the ground-truth bounding box for each object. This protocol evaluates both object geometry and its placement within the composed scene.

\subsection{Canonical-Space Single-Object Generation}
\label{sec:exp:canon}

\paragraph{Setup and baselines.}
We evaluate on a fixed set of $500$ randomly selected objects from Toys4k~\cite{stojanov2021using}. We compare against the official Pixal3D~\cite{li2026pixal3d} pipeline, whose released model supports only single-view input, TRELLIS~\cite{xiang2025structured} in both its single-image and multi-image modes, and TRELLIS.2~\cite{xiang2025trellis2}, whose released model also supports only a single input view. Our model is evaluated with $1$--$16$ views. All predictions are decoded and aligned using the same evaluation protocol. Table~\ref{tab:main} reports generation quality across different numbers of input views ($1,2,4,8,16$) and per-view occlusion fractions ($0,25,50,75\%$).

\paragraph{Results.}
On clean inputs, our model is competitive with the official single-view Pixal3D pipeline when using one view and outperforms the baselines as additional views become available. Performance improves consistently with the number of input views. Since our model is trained with a variable number of views rather than optimized specifically for single-image generation, matching a dedicated single-view prior in the one-view setting indicates that the multi-view adaptation largely preserves the original generative capability while enabling substantial gains from additional observations.

The advantage becomes more pronounced under occlusion. As the number of views increases, our model remains robust even when individual observations are heavily incomplete. With $16$ views, performance changes only slightly up to $50\%$ per-view occlusion, whereas both the single-view baselines and the single-view variant of our model degrade substantially. Across all metrics, additional views considerably reduce the effect of occlusion, with distance errors remaining stable and F-Scores decreasing only mildly in the high-view regime. These results show that multi-view conditioning, together with the conditioning-view augmentation, provides strong robustness to partial observations.

\subsection{Multi-object Scene Generation}
\label{sec:exp:multi}

\paragraph{Setup and baselines.}
We evaluate compositional scene generation on both synthetic and real-world multi-object scenes. For Toys4k-Scene, we construct cluttered scenes from Toys4k~\cite{stojanov2021using} objects and render each scene along a short orbiting camera trajectory. We additionally evaluate on HouseCat6D~\cite{jung2024housecat6d}, which provides real-world captures together with per-object ground-truth meshes. For both datasets, each predicted object is placed into the shared world frame and evaluated directly against its corresponding ground-truth geometry.

We compare against a representative set of compositional and generative scene reconstruction methods, including SAM3D~\cite{chen2026sam}, SceneGen~\cite{meng2026scenegen}, SceneMaker~\cite{shi2026scenemaker}, RecGen~\cite{zadaianchuk2026recgen}, and ShapeR~\cite{siddiqui2026shaperrobustconditional3d}. Methods that require depth additionally receive ground-truth depth maps. We use the same evaluation protocol for UE-MeshyScene.

\begin{figure}[t]\centering
\includegraphics[width=\linewidth]{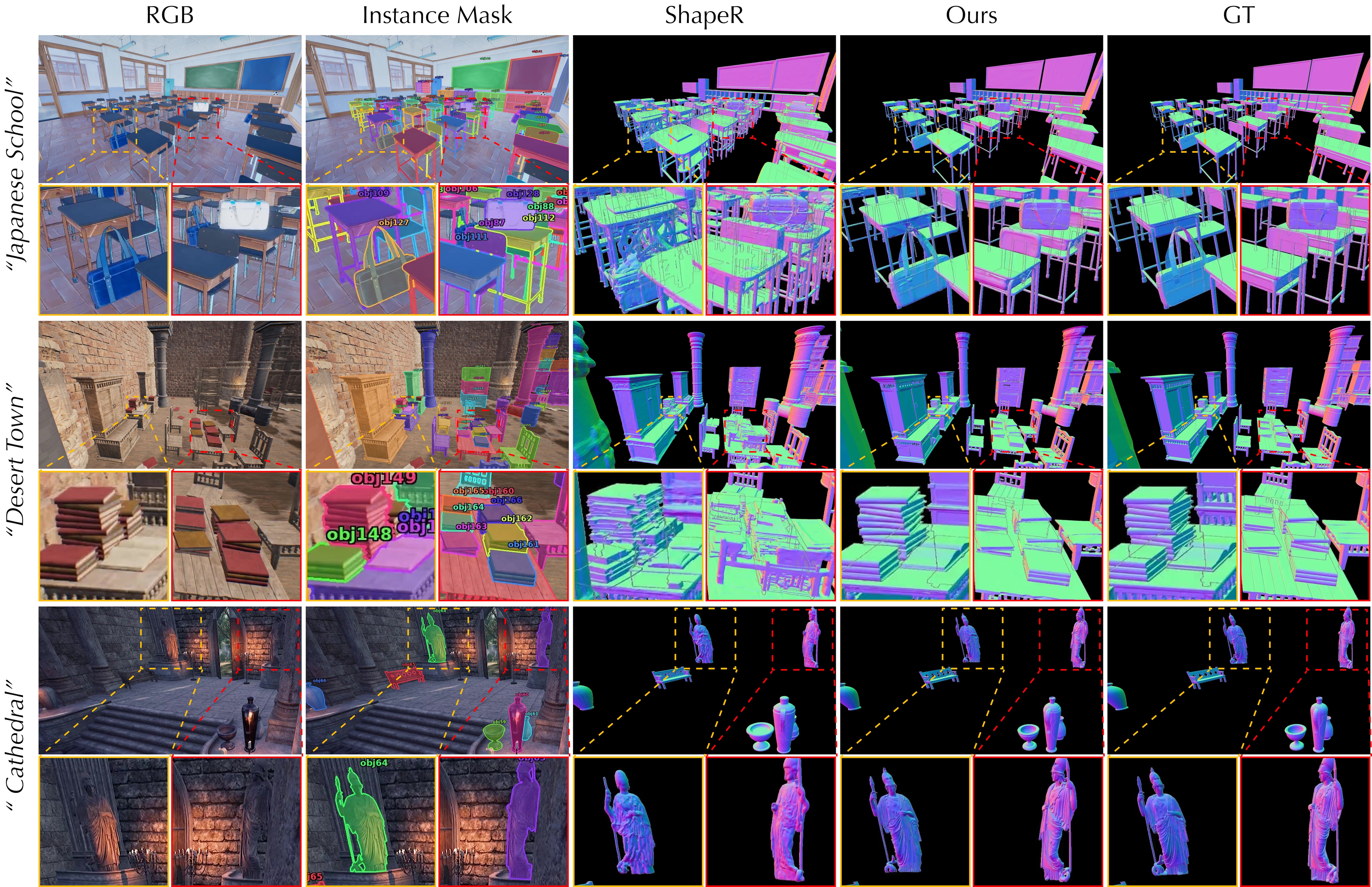}
\setlength{\abovecaptionskip}{-2mm}
\caption{\textbf{Qualitative comparison on UE-MeshyScene.}
We compare per-object meshes generated in densely cluttered and heavily occluded scenes against the multi-view baseline ShapeR~\cite{siddiqui2026shaperrobustconditional3d}. All predicted meshes are placed in the shared world frame and visualized using surface normals. Our method recovers more detailed and accurately placed geometry, including for small and severely occluded objects, while the baseline often produces over-smooth or misplaced shapes in highly cluttered regions. These qualitative differences are consistent with the quantitative results reported in Table~\ref{tab:ue}.}
\label{fig:ue_scene}
\end{figure}

\paragraph{Qualitative comparison.}
Figures~\ref{fig:toys4k_scene} and~\ref{fig:housecat6d} compare the methods qualitatively on Toys4k-Scene and HouseCat6D, respectively. Alongside renderings of the composed scenes, we visualize per-pixel geometric discrepancies within selected zoomed-in regions. For pixels where the predicted and ground-truth surfaces overlap, color indicates the angular error between their surface normals, ranging from blue at \(0^\circ\) to red at \(90^\circ\). Non-overlapping regions are visualized separately: pink denotes spurious predicted surfaces, while gray denotes ground-truth surfaces that are missing from the prediction. These regions also contribute to the mask-IoU error.

\subsection{UE-MeshyScene Generation}
\label{sec:exp:ue}

\paragraph{Setup.}
We evaluate our method on the six UE-MeshyScene environments introduced above. This benchmark provides a substantially more challenging setting than previous dataset, with hundreds of objects, severe mutual occlusion, and a broader range of viewpoints. In such densely cluttered scenes, a single image often reveals only a small fraction of many objects. We therefore focus our comparison on ShapeR~\cite{siddiqui2026shaperrobustconditional3d}, a multi-view baseline that can incorporate observations across multiple viewpoints. Both methods are evaluated using the same world-frame protocol described above, without per-object alignment.

\paragraph{Results.}
Table~\ref{tab:ue} reports the quantitative results on UE-MeshyScene. Our method outperforms ShapeR across all reported metrics. The improvement is also more pronounced in the median than in the mean, indicating that the performance gap is not driven by a small number of difficult objects but is consistently observed across typical objects in the scenes. Figure~\ref{fig:ue_scene} shows qualitative comparisons. Our method recovers more detailed object geometry while maintaining more accurate placement in the shared world frame, particularly for small and heavily occluded objects.

\subsection{Ablation: Multi-View Fusion}
\label{sec:exp:ablation}

\paragraph{Setup.}
We ablate the multi-view fusion module by replacing the learned aggregator in Eq.~\eqref{eq:ibr} with a simple arithmetic mean over the per-view 3D feature grids (``Avg''). Both variants start from the same fused representation: since the final layer of the IBR feature MLP is zero-initialized, its residual branch initially contributes nothing, and the learned aggregator reduces exactly to the cross-view mean. During training, the learned variant can move beyond this initialization by assigning voxel-wise, view-dependent residual contributions. We follow the same canonical-space evaluation protocol on Toys4k as in Sec.~\ref{sec:exp:canon} and Table~\ref{tab:main}, using the same $500$ objects, input-view counts, and per-view occlusion levels. The fusion module is the only component changed between the two variants. Results are reported in Table~\ref{tab:ablation}.

\begin{table*}[t]\centering
\setlength{\abovecaptionskip}{2mm}
\caption{\textbf{Fusion ablation} on the Toys4k~\cite{stojanov2021using} canonical-space setting of Table~\ref{tab:main}. We compare the learned IBR aggregator with arithmetic mean fusion (Avg) of the per-view feature grids under varying numbers of input views (\#V) and per-view occlusion levels. We report CD-$\ell_2$ ($\times10^{-3}$, $\downarrow$), CD-$\ell_1$ ($\times10^{-2}$, $\downarrow$), EMD ($\times10^{-2}$, $\downarrow$), and F-Score@$\{.02,.05\}$ ($\uparrow$). The better fusion strategy for each setting and metric is shown in \textbf{bold}.}
\label{tab:ablation}
\setlength{\tabcolsep}{3pt}
\resizebox{\linewidth}{!}{%
\begin{tabular}{c l *{20}{c}}
\toprule
 & & \multicolumn{5}{c}{Occl.\ $0\%$} & \multicolumn{5}{c}{Occl.\ $25\%$}
   & \multicolumn{5}{c}{Occl.\ $50\%$} & \multicolumn{5}{c}{Occl.\ $75\%$} \\
\cmidrule(lr){3-7}\cmidrule(lr){8-12}\cmidrule(lr){13-17}\cmidrule(lr){18-22}
\#V & Fusion
 & CD-$\ell_2$ & CD-$\ell_1$ & EMD & F@.02 & F@.05
 & CD-$\ell_2$ & CD-$\ell_1$ & EMD & F@.02 & F@.05
 & CD-$\ell_2$ & CD-$\ell_1$ & EMD & F@.02 & F@.05
 & CD-$\ell_2$ & CD-$\ell_1$ & EMD & F@.02 & F@.05 \\
\midrule
\multirow{2}{*}{1} & Avg
 & 2.83 & 3.96 & \textbf{7.7} & 0.731 & 0.927
 & 4.07 & 4.83 & \textbf{8.7} & 0.666 & 0.893
 & 9.53 & 7.03 & \textbf{11.2} & \textbf{0.557} & 0.812
 & \textbf{25.43} & 12.35 & 16.9 & \textbf{0.372} & \textbf{0.633} \\
 & IBR
 & \textbf{2.59} & \textbf{3.85} & 7.8 & \textbf{0.739} & \textbf{0.932}
 & \textbf{3.70} & \textbf{4.73} & \textbf{8.7} & \textbf{0.669} & \textbf{0.896}
 & \textbf{8.90} & \textbf{6.84} & \textbf{11.2} & 0.554 & \textbf{0.818}
 & 25.45 & \textbf{12.30} & \textbf{16.8} & 0.370 & 0.631 \\
\midrule
\multirow{2}{*}{2} & Avg
 & 1.58 & 3.33 & \textbf{7.2} & 0.778 & 0.951
 & 2.65 & 3.95 & \textbf{7.9} & 0.723 & 0.930
 & 4.30 & 5.05 & 8.9 & 0.642 & 0.884
 & 12.91 & 8.72 & 13.2 & 0.471 & 0.742 \\
 & IBR
 & \textbf{1.54} & \textbf{3.23} & \textbf{7.2} & \textbf{0.790} & \textbf{0.954}
 & \textbf{2.57} & \textbf{3.87} & \textbf{7.9} & \textbf{0.734} & \textbf{0.931}
 & \textbf{4.01} & \textbf{4.85} & \textbf{8.8} & \textbf{0.651} & \textbf{0.893}
 & \textbf{10.03} & \textbf{7.96} & \textbf{12.1} & \textbf{0.494} & \textbf{0.762} \\
\midrule
\multirow{2}{*}{4} & Avg
 & 1.43 & 3.17 & 7.1 & 0.796 & 0.958
 & 2.00 & 3.52 & \textbf{7.3} & 0.760 & 0.947
 & 2.71 & 4.13 & 7.9 & 0.703 & 0.923
 & 6.10 & 6.04 & 10.2 & 0.579 & 0.845 \\
 & IBR
 & \textbf{1.35} & \textbf{3.02} & \textbf{7.0} & \textbf{0.813} & \textbf{0.964}
 & \textbf{1.62} & \textbf{3.35} & \textbf{7.3} & \textbf{0.778} & \textbf{0.950}
 & \textbf{2.56} & \textbf{4.01} & \textbf{7.8} & \textbf{0.719} & \textbf{0.925}
 & \textbf{5.67} & \textbf{5.86} & \textbf{10.1} & \textbf{0.590} & \textbf{0.849} \\
\midrule
\multirow{2}{*}{8} & Avg
 & 1.55 & 3.13 & 7.1 & 0.806 & 0.960
 & 1.51 & 3.26 & \textbf{7.2} & 0.783 & \textbf{0.957}
 & \textbf{1.89} & 3.63 & \textbf{7.5} & 0.744 & 0.942
 & \textbf{3.04} & \textbf{4.65} & \textbf{8.6} & 0.650 & 0.898 \\
 & IBR
 & \textbf{1.23} & \textbf{2.94} & \textbf{6.9} & \textbf{0.822} & \textbf{0.965}
 & \textbf{1.50} & \textbf{3.20} & \textbf{7.2} & \textbf{0.796} & 0.956
 & 1.91 & \textbf{3.57} & \textbf{7.5} & \textbf{0.755} & \textbf{0.943}
 & 3.52 & \textbf{4.65} & \textbf{8.6} & \textbf{0.663} & \textbf{0.902} \\
\midrule
\multirow{2}{*}{16} & Avg
 & 1.30 & 3.02 & \textbf{6.9} & 0.812 & 0.964
 & 1.37 & 3.11 & 7.1 & 0.800 & 0.962
 & 1.57 & 3.33 & \textbf{7.2} & 0.775 & 0.954
 & 2.43 & 4.13 & 8.1 & 0.695 & 0.922 \\
 & IBR
 & \textbf{1.19} & \textbf{2.89} & \textbf{6.9} & \textbf{0.827} & \textbf{0.967}
 & \textbf{1.29} & \textbf{3.01} & \textbf{7.0} & \textbf{0.813} & \textbf{0.964}
 & \textbf{1.44} & \textbf{3.23} & \textbf{7.2} & \textbf{0.788} & \textbf{0.955}
 & \textbf{2.25} & \textbf{3.97} & \textbf{8.0} & \textbf{0.713} & \textbf{0.929} \\
\bottomrule
\end{tabular}
}
\end{table*}

\begin{table}[t]\centering
\setlength{\abovecaptionskip}{2mm}
\caption{\textbf{Scene-level fusion ablation.}
We compare the learned IBR aggregator with arithmetic mean fusion (Avg) on the compositional benchmarks in Table~\ref{tab:scene} and UE-MeshyScene in Table~\ref{tab:ue}. All methods are evaluated using the same world-frame, no-ICP protocol, with F-Score computed at $\tau{=}0.05$. We report CD-$\ell_2$ ($\times10^{-3}$, $\downarrow$), CD-$\ell_1$ ($\times10^{-2}$, $\downarrow$), and EMD ($\times10^{-2}$, $\downarrow$). The best result in each column is shown in \textbf{bold}.}
\label{tab:ablation-scene}
\setlength{\tabcolsep}{6pt}
\resizebox{\linewidth}{!}{%
\begin{tabular}{l *{12}{c}}
\toprule
Fusion & \multicolumn{4}{c}{HouseCat6D~\cite{jung2024housecat6d}}
   & \multicolumn{4}{c}{Toys4k-Scene~\cite{stojanov2021using}}
   & \multicolumn{4}{c}{UE-MeshyScene} \\
\cmidrule(lr){2-5}\cmidrule(lr){6-9}\cmidrule(lr){10-13}
 & CD-$\ell_2$$\downarrow$ & CD-$\ell_1$$\downarrow$ & EMD$\downarrow$ & F-Score$\uparrow$
 & CD-$\ell_2$$\downarrow$ & CD-$\ell_1$$\downarrow$ & EMD$\downarrow$ & F-Score$\uparrow$
 & CD-$\ell_2$$\downarrow$ & CD-$\ell_1$$\downarrow$ & EMD$\downarrow$ & F-Score$\uparrow$ \\
\midrule
Avg
 & \textbf{0.27} & 1.60 & \textbf{3.3} & \textbf{0.996}
 & 0.65 & 1.88 & 3.8 & \textbf{0.981}
 & 2.83 & 2.92 & 5.1 & 0.944 \\
IBR
 & 0.28 & \textbf{1.53} & 3.4 & 0.995
 & \textbf{0.61} & \textbf{1.78} & \textbf{3.7} & \textbf{0.981}
 & \textbf{2.48} & \textbf{2.66} & \textbf{5.0} & \textbf{0.951} \\
\bottomrule
\end{tabular}
}
\end{table}

\paragraph{Results.}
The learned IBR aggregator performs comparably to or better than mean averaging in most settings, with its advantage becoming more evident as the number of input views and the occlusion level increase. With a single input view, the two variants perform similarly, since there are no additional observations to reweight. Once multiple views are available, IBR shows a consistent benefit. For example, with $2$ views and $75\%$ occlusion, IBR reduces CD-$\ell_2$ from $12.91$ to $10.03$ and improves F@.05 from $0.742$ to $0.762$. With $16$ views, it outperforms mean fusion on nearly every metric across all occlusion levels. This trend suggests that learned view weighting is most useful when multiple observations provide features of varying reliability. The few settings in which mean averaging performs slightly better, primarily at $8$ views, show only small differences. We therefore use the learned IBR aggregator as the default multi-view fusion module.

\paragraph{Scene-level ablation.}
We further compare the two fusion modules end-to-end in the compositional setting using the same world-frame, no-ICP protocol as Table~\ref{tab:scene}. The three benchmarks form a natural progression in scene complexity, from the relatively sparse and mildly occluded HouseCat6D scenes, to the denser Toys4k-Scene layouts, and finally to the heavily cluttered UE-MeshyScene benchmark with hundreds of mutually occluding objects. As shown in Table~\ref{tab:ablation-scene}, the benefit of learned IBR aggregation becomes increasingly pronounced along this progression. On HouseCat6D, the two fusion strategies perform similarly, suggesting that learned view weighting offers limited benefit when objects are well observed. IBR consistently improves the distance metrics on the more cluttered Toys4k-Scene and achieves its largest gains on UE-MeshyScene, where it improves all reported metrics over mean fusion. For example, on UE-MeshyScene, IBR reduces CD-$\ell_2$ by $12\%$ and increases F-Score from $0.944$ to $0.951$. This trend is consistent with the canonical-space ablation in Table~\ref{tab:ablation}: learned aggregation becomes most useful when multiple views provide observations of varying reliability, as commonly occurs in densely cluttered and heavily occluded scenes.

\begin{figure}[t]\centering
\includegraphics[width=\linewidth]{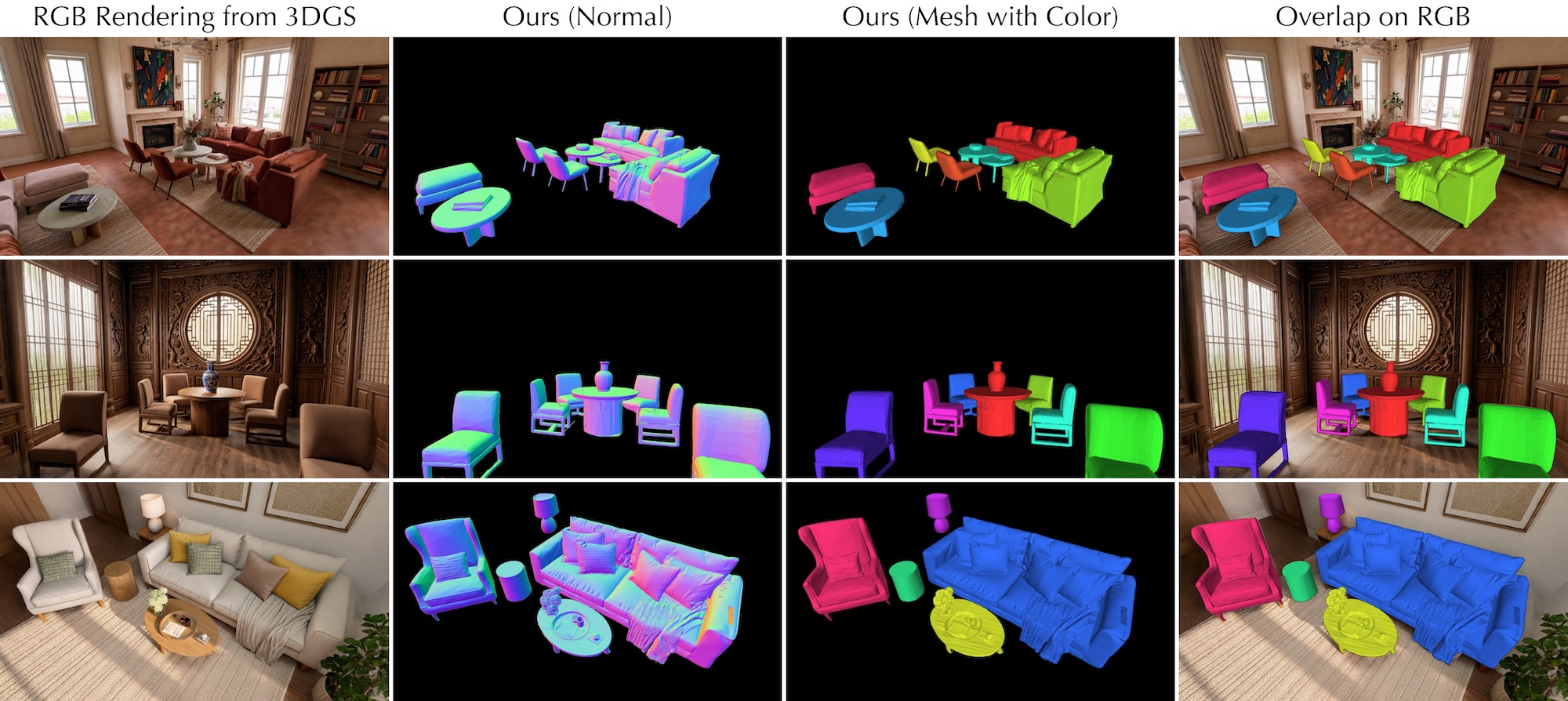}
\setlength{\abovecaptionskip}{-2mm}
\caption{\textbf{Application to a Marble-generated 3DGS world.}
Given a generated 3DGS scene, we render a sequence of posed observations and apply WorldSculpt to convert the scene into a compositional representation of individual per-object meshes in a shared world frame.}
\label{fig:marble}
\end{figure}

\subsection{Application: Converting Generated 3D Worlds into Compositional Meshes}
\label{sec:exp:marble}

Generative world models such as Marble~\cite{worldlabs2025marble} and HY-World 2.0~\cite{team2026hy} can synthesize photorealistic 3D environments represented as 3D Gaussian Splatting (3DGS). While such representations support high-quality free-viewpoint rendering, they do not explicitly separate the scene into editable object-level meshes required by downstream authoring, simulation, and asset reuse. As a proof of concept, we apply WorldSculpt to a world generated by Marble and convert its 3DGS representation into a compositional mesh scene without any additional training.

\noindent\textbf{(i) Render posed observations from the 3DGS.}
We obtain the reconstructed 3DGS of a Marble scene and render a sequence of RGB frames along a virtual camera trajectory. For each frame, we additionally render a pixel-aligned metric depth map from the same Gaussian representation using the expected-depth renderer in gsplat. Since both RGB and depth are rendered with a controlled virtual camera in the coordinate system of the 3DGS, the camera intrinsics and poses are known exactly, and the depth maps and gravity direction are expressed in the same world frame.

\noindent\textbf{(ii) Recover object masks and coarse 3D localizations.}
We apply SAM3~\cite{carion2025sam} to the rendered sequence to obtain per-object mask tracks. A physical object may occasionally be split into multiple tracklets because of duplicate detections or temporary tracking failures. We merge such fragments using both temporal mask overlap and 3D consistency. Tracklets that coexist with high mask IoU are first merged in 2D, after which same-category tracklets with strongly overlapping world-space boxes are merged when they occur in disjoint temporal intervals. Object boxes are re-estimated after each merge until the tracks stabilize.

For each consolidated track, we estimate a coarse 3D localization from its masks and rendered depth. Valid masked pixels are back-projected through the known camera intrinsics and poses to form partial world-space point clouds. We discard frames with outlying point-cloud centroids, merge the remaining observations, and retain the dominant cluster using DBSCAN~\cite{ester1996density}. An axis-aligned bounding box of the resulting point cloud is then used as the coarse localization required by WorldSculpt.

\noindent\textbf{(iii) Apply WorldSculpt.}
The resulting posed RGB frames, per-object mask tracks, and coarse 3D localization boxes match the input assumptions of Sec.~\ref{sec:method:test}. We therefore apply the standard WorldSculpt inference pipeline without modification to generate a collection of individual per-object meshes in the shared world frame, as shown in Figure~\ref{fig:marble}. This example demonstrates that a generated 3DGS world can be converted into a compositional mesh representation using the same object-generation model, without scene- or domain-specific retraining.

\subsection{Limitations}
\label{sec:exp:limitations}

WorldSculpt has several limitations. First, our method relies on upstream estimates of camera poses, per-object instance masks, and coarse 3D localization boxes. The conditioning-view augmentation curriculum in Sec.~\ref{sec:method:train} improves robustness to moderate errors in these inputs, but large inaccuracies, such as severely misplaced boxes or erroneous masks, can still affect the generated geometry. A more integrated system that jointly performs object detection, localization, and generation could reduce this dependence on upstream predictions. Performance may also deteriorate under challenging capture conditions that are not explicitly modeled during training, including low-light degradation~\cite{niu2023visibility,niu2026robust3dawarergbnirimaging,niu2023nir,niu2023physics} and motion-induced artifacts~\cite{ji2023rethinking,ji2026moment,kupyn2019deblurgan,niu2024bundle,niu2024rs}.

Second, the current model generates geometry only. We build on the first two geometry stages of Pixal3D and do not model object appearance. Both Pixal3D~\cite{li2026pixal3d} and TRELLIS.2~\cite{xiang2025trellis2} support physically based rendering materials, making it possible to extend our multi-view conditioning pathway to their texture and material generation stages. This would allow compositional scenes to be generated with both detailed geometry and consistent object appearance.

Third, our formulation assumes a static scene. Multi-view observations of each object are treated as measurements of the same fixed geometry and pose, so moving or deforming objects are not currently supported. Extending WorldSculpt to dynamic scenes would require estimating object motion and accounting for non-rigid changes across observations. Possible directions include incorporating optical-flow-based motion estimation~\cite{sun2018pwc,teed2020raft,niu2024mofa,ilg2017flownet} and introducing articulated representations such as skeletal models~\cite{yang2023effective} or SMPL-X-based models~\cite{pavlakos2019expressive,loper2023smpl,niu2026motion,zhan2025towards,niu2025anicrafter} for objects undergoing articulated deformation.

\section{Conclusion}
\label{sec:conclusion}

We presented \emph{WorldSculpt}, a framework for generating compositional 3D scenes from grounded multi-view observations. WorldSculpt adapts a strong object-level generative prior with a multi-view conditioning pathway that integrates observations of each object in an anchor-aligned canonical frame. The generative prior provides plausible completion of occluded and unobserved geometry, while multi-view conditioning grounds the generated shape in the available image evidence. Although the model is trained entirely on individual objects in canonical space, it generalizes without scene-level retraining to large, densely cluttered scenes at inference time.

To support evaluation in this setting, we introduced UE-MeshyScene, a photorealistic synthetic benchmark containing six scenes with $93$ to $701$ objects each. Across controlled single-object evaluations, multi-object scene benchmarks, and large-scale UE-MeshyScene experiments, WorldSculpt consistently outperforms the evaluated baselines, with the largest gains appearing under dense clutter and severe occlusion. Our fusion ablations show a similar trend, with learned multi-view aggregation becoming increasingly beneficial as observations grow more incomplete and heterogeneous.

Future work could extend the same conditioning mechanism to texture and material generation for joint geometry and appearance synthesis. Another important direction is to relax the static-scene assumption and support moving or deforming objects. Overall, our results show that adapting object-level generative priors to multi-view compositional generation provides a practical path toward editable 3D representations of complex, cluttered environments.

\bibliographystyle{abbrv}
\bibliography{references}

\end{document}